\documentclass{article} 
\usepackage{iclr2017_conference,times}
\usepackage{hyperref}
\usepackage{url}
\usepackage{graphicx}
\usepackage{amsmath}
\usepackage{amsfonts}

\title{Accelerating SGD for Distributed Deep-Learning Using Approximted Hessian Matrixt }

\author{S\'{e}bastien Arnold \\
University of Southern California\\
Los Angeles, CA-90007, USA \\
\texttt{arnolds@usc.edu} \\
\And
Chunming Wang \\
University of Southern California\\
Los Angeles, CA-90007, USA \\
\texttt{cwang@math.usc.edu} \\
}

\begin{document}

\maketitle

\begin{abstract}
We introduce a novel method to compute a rank $m$ approximation of the inverse of the Hessian matrix in the distributed regime. By leveraging the differences in gradients and parameters of multiple Workers, we are able to efficiently implement a distributed approximation of the Newton-Raphson method. We also present preliminary results which underline advantages and challenges of second-order methods for large stochastic optimization problems. In particular, our work suggests that novel strategies for combining gradients provide further information on the loss surface.

\end{abstract}

\section{Introduction}
\label{introduction}
The Stochastic Gradient Descent (SGD) method has been shown to be well-suited for distributed deep-learning applications \citep{Mitliagkas, rudra, downpour, deepspeech}. Typically a set of Worker nodes evaluate the gradient of the cost functional on one or a batch of data point and perform updates to the parameters acquired (Read) from a Parameter Server in parallel either synchronously or asynchronously. Periodically, the Workers return (Write) updated parameters to the Server. In most of cases, the average of the returned parameter values is evaluated by the Server and is made available to Workers for the subsequent acquisitions. In some implementations, the Workers are not required to return the gradient vectors of the cost functional to the server. In our method, the Workers supply the gradient evaluated at the updated parameters to the Server. Using these gradient vectors the Server uses an approximated Hessian matrix to produce a quasi-Newton update of the parameter which is made available to Workers. Our numerical experimental results show that the new approach leads to accelerated convergence of the parameter, as well as, reduction of the cost functional. In some cases, the new algorithm exhibits quadratic convergence of the parameter which is characteristically associated with the Newton's method.


\section{Method}
\label{theory}
Following each Write operation by a Worker or all Workers in either an asynchronous or synchronous implementation of distributed SGD algorithm, the Server receives the updated parameters \(\theta_k\in\mathbb{R}^n\) and the estimated gradient \(\bigtriangledown J(\theta_k)\) from Workers \(k=1,\cdots,m\). An approximation of the Hessian matrix \(H_J\) can be obtained by requiring the following equality:
\begin{equation}\label{eq:Hessian}
\bigtriangledown J(\theta_k)-\bigtriangledown J(\theta_j)=H_J(\theta_k-\theta_j),\forall k,j=1,\cdots,m.
\end{equation}
We define \(n\times m\) matrices \(G\) and \(\Theta\) such that the \(k\)-th columns of these matrices are \(\bigtriangledown J(\theta_k)-\bar{g}\) and \(\theta_k-\bar{\theta}\), respectively, where
\begin{displaymath}
\bar{g}=\frac{1}{m}\sum^m_{k=1}\bigtriangledown J(\theta_k),\qquad \bar{\theta}=\frac{1}{m}\sum^m_{k=1}\theta_k.
\end{displaymath}
Equation (\ref{eq:Hessian}) leads to \(G=H_J\Theta\) which represents key characteristics of the Hessian matrix. Note that both matrices \(G\) and \(\Theta\) are not square matrices and not invertible in general. Therefore the above equality does not uniquely define the matrix \(H_J\). Our objective is to find a rank \(p\) approximation \(\bar{H}^{-1}_J\) of the inverse of the Hessian matrix and to generate an update to the parameter
\begin{equation}\label{eq:Newton}
\theta_{new}=\bar{\theta}-\tau\bar{H}^{-1}_J\bar{g}.
\end{equation}
Our selection for \(\bar{H}^{-1}_J\) is guided by the following observations. Consider a singular value decomposition of the matrix \(G\) given by \(U_G\Sigma_G V^H_G\) where the first \(m\) columns of the unitary matrix 
\(V_G\) are eigenvectors of the matrix \(G^HG\) and \(\Sigma_G\) is a \(n\times m\) diagonal matrix with 
\(\sigma_{1,1}\geq\sigma_{2,2}\geq\cdots\sigma_m\geq 0\). Note that the first \(m\) columns of the matrix 
\(U_G\) are given by \(u_k=Gv_k/\|Gv_k\|_2,k=1,\cdots,m\) where \(v_k\) is the \(k\)-th column of the matrix \(V_G\). Equation \(G=H_J\Theta\) can be rewritten as \(H^{-1}_JU_G\Sigma_G=\Theta V_G\). We define \(n\times n\) matrix \(P=H^{-1}_JU_G\) and \(n\times m\) matrix \(Y=\Theta V_G\), by matching the first 
\(m\) columns we obtain \(p_k=\sigma^{-1}_k y_k\) where \(p_k\) and \(y_k=\Theta v_k\) are the \(k\)-th columns of matrices \(P\) and \(Y\), respectively. We define an approximation of the matrix \(H^{-1}_J\) by
\begin{equation}\label{eq:approxHessian}
\bar{H}^{-1}_J z=\left\{\begin{array}{cc}
PU^H_Gz, & z\in span\{u_k,k=1,\cdots,j\},\\
z,& z\in span\{u_k,k=1,\cdots,j\}^\bot,
\end{array}\right.
\end{equation}
where the integer \(j\) is satisfies \(\sigma_j\geq\lambda\sigma_1,\sigma_{j+1}<\lambda\sigma_1\) for a selected value \(0<\lambda<1\). In particular, for any \(z\in\mathbb{R}^n\) such that
\begin{displaymath}
z=\sum^j_{k=1}\alpha_k u_k +z^\bot,\qquad \alpha_k=z^Hu_k, k=1,\cdots,j,
\end{displaymath}
we define
\begin{equation}\label{eq:HessianInverse}
\bar{H}^{-1}_J z=z^\bot+\sum^j_{k=1}\alpha_k\sigma^{-1}_k y_k. 
\end{equation}
Note that when \(j=0\), equation (\ref{eq:Newton}) is equivalent to the standard SGD method. The algorithm does require computation of the eigenvalues and eigenvectors of the \(m\times m\) matrix \(G^HG\). For small enough number of Workers, this represents a relative minor computational effort. 

We underline that while this presentation follows the parameter server \citep{param-server} semantics, our technique is easily adapted to the tree-reduction framework \citep{firecaffe}. In fact, it can be seen as a more sophisticated reduction of the gradients across Workers, as opposed to a simple averaging. Our MPI-based implementation consists of a large All-to-All broadcast of the parameters and gradients, followed by the computations presented above. With $n$ parameters and $m$ replicas, this method has a space complexity on the order of $\mathcal{O}(mn)$ and its time complexity is $\mathcal{O}(m^3 + m \cdot n)$.


\section{Experimental Results}
\label{results}

\begin{figure}[h]
\begin{center}
\includegraphics[width=1.0\textwidth]{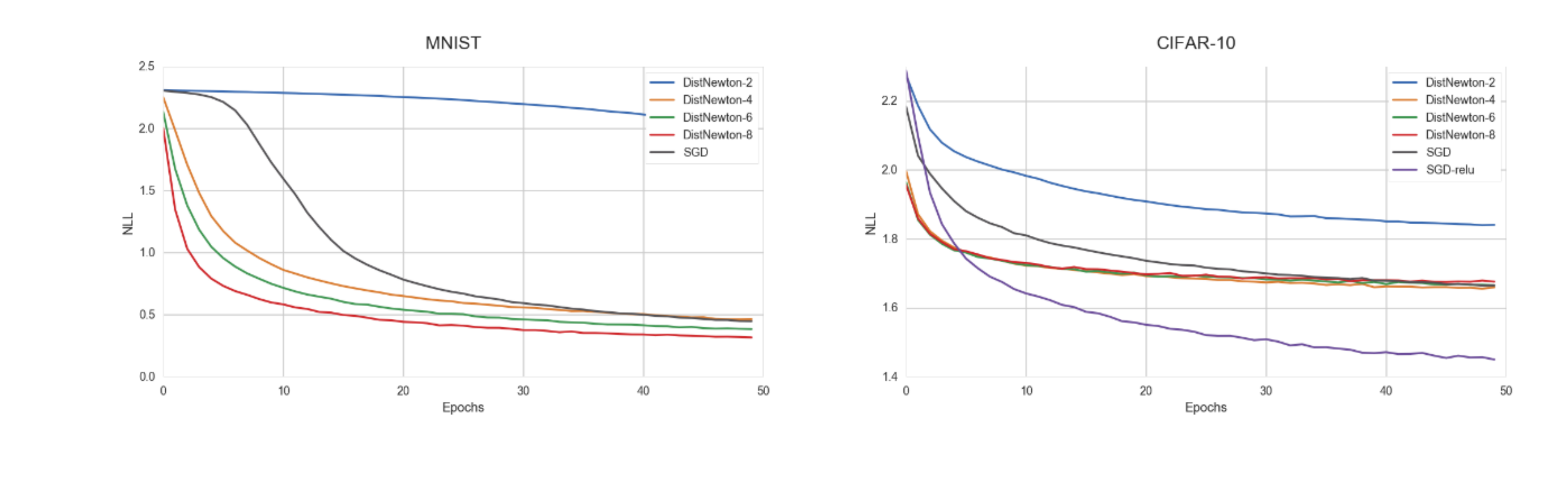}
\end{center}
\caption{Convergence curves on MNIST and CIFAR-10. DistNewton-$m$ denotes the use of $m$ Workers. In both experiments we notice an improvement in convergence as $m$ increases. (ie, more gradients are used to compute $\bar{H}^{-1}$). On the CIFAR-10 experiment we also plot the SGD ReLU performance, which outperforms tanh activations and diverged using our method.}
\end{figure}

We now present preliminary results on two widely used datasets (MNIST \citep{mnist} and CIFAR-10 \citep{cifar10}) and compare the convergence rate of our proposed method against SGD. Furthermore, we restrict our study to the synchronous case. Since we are only interested in the optimization performance, we keep most of our hyper-parameters constant, including learning rates (0.0003 and 0.01), activation functions (ReLU \citep{relu} and tanh), and a global batch size of 256. Our model is a 5 layer convnet with about 16'000 parameters, which we train each time for 50 epochs. We report the negative log-likelihood on the train dataset at every epoch, and for up to 8 Workers. Note that since the global batch size is fixed, the SGD convergence curves are identical and thus only reported once.

Our results clearly demonstrate convergence improvements as we scale to a larger number of Workers, and consistently outperforms stochastic gradient descent in the most distributed case. Interestingly the latter is true even with a relatively small number of Workers, as we observe a much faster convergence with $m = 4$ in both experiments. However, when the number of Workers is not sufficient to properly estimate the most influential singular values the method converges to poor minimas and is slower than distributed SGD. This effect underlines the importance of the number of eigenvalues considered which is defined through the parameter $j$.

Additionally, our method suffers of the limitations of Newton's method. For example, several experiments diverged when using too large a learning rate, whereas this was beneficial to the convergence rate of SGD. Another downside is related to the use of ReLU activations; a good enough estimate of the Hessian results in numerical errors as the second derivative becomes 0. However, ReLUs have been widely successful in the computer vision domain \citep{alexnet,resnet}, and usually outperform other non-linear activations. This is demonstrated by the SGD-relu curve in the CIFAR-10 experiment. Finally, as pointed by \cite{dauphin-saddle}, even when including second-order information, iterative methods such as SGD or Newton method can be slowed down by saddle-points surrounded by plateaus. 

We note that previous work has suggested approaches to tackle some of those limitations. In particular, \cite{lecun-lr} derived an optimal formulation of the learning rate, assuming knowledge of the largest singular value $\sigma_{\text{max}}$: $\tau_{\text{opt}} = \frac{1}{\sigma_{\text{max}}}$. Since we directly approximate $\sigma_{\text{max}}$ we can trivially adapt the learning rate to be upper-bounded by this approximate optimal at every update. In addition to underlining the saddle-point issue, \cite{dauphin-saddle} also proposed a counter-measure: considering the absolute value of the Hessian's singular values. This approach comes as an artifact of our suggested method, since we approximate the inverse of the Hessian using only positive singular values.



\section{Conclusion}
\label{Conclusion}

In this work, we introduced a Quasi-Newton method specifically designed for the distributed regime. On preliminary small-scale experiments, our method largely outperforms stochastic gradient descent when the number of Workers allow for a good approximation of the inverse of the Hessian. Our results suggest that our method is effectively taking advantage of the second-order information of the optimization problem.

More importantly, this work suggests that alternative strategies for combining Workers' gradients will provide superior convergence rates than a simple averaging. Intuitively this results from the observation that each Worker is exploring a different region of the loss surface and thus the aggregated information will provide a better understanding than averaged local statistics.

Finally, we would like to indicate that our work is preliminary and more comprehensive investigation into the potential of this approach is required. In particular, we want to further define criteria to identify cases for which this approach offers maximum benefits. 

\subsubsection*{Acknowledgments}
Sponsorship of the Living With a Star Targeted Research and Technology NASA/NSF
Partnership for Collaborative Space Weather Modeling is gratefully acknowledged.

\bibliographystyle{plainnat}
\bibliography{biblio}


\end{document}